%% file: main.tex
\documentclass[10pt,twocolumn,letterpaper]{article}

\usepackage[pagenumbers]{cvpr} 

\input{preamble}

\definecolor{cvprblue}{rgb}{0.21,0.49,0.74}
\usepackage[pagebackref,breaklinks,colorlinks,allcolors=cvprblue]{hyperref}

\def\paperID{}
\def\confName{CVPR}
\def\confYear{2026}

\newcommand{\ours}{SoliReward}

\usepackage{caption}
\title{{\ours}: Mitigating Susceptibility to Reward Hacking and \\ Annotation Noise in Video Generation Reward Models}

\author{
Jiesong Lian$^{1,\dagger,*}$ \quad 
Ruizhe Zhong$^{2,\dagger,*}$ \quad
Zixiang Zhou$^{3}$ \quad
Xiaoyue Mi$^{4}$ \quad \\
Long Hu$^{1,\S}$ \quad
Yuan Zhou$^{3,\ddagger}$ \quad
Qinglin Lu$^{3}$ \quad
Yixue Hao$^{1}$ \quad
Junchi Yan$^{2,\S}$ \\
$^1$Huazhong University of Science and Technology \quad
$^2$Shanghai Jiao Tong University \quad \\
$^3$Tencent Hunyuan \quad
$^4$University of Chinese Academy of Sciences}

\begin{document}
\maketitle

\footnotetext[1]{$\dagger$ Work done during internship at Tencent Hunyuan.}
\footnotetext[2]{$*$ Equal contribution.
\\ \hspace*{2.3em} lian700@hust.edu.cn, zerzerzerz271828@sjtu.edu.cn}
\footnotetext[3]{
$\ddagger$ Project leader.}
\footnotetext[4]{$\S$ Corresponding author. hulong@hust.edu.cn, yanjunchi@sjtu.edu.cn}

\begin{abstract}
Post-training alignment of video generation models with human preferences is a critical goal. Developing effective Reward Models (RMs) for this process faces significant methodological hurdles.
Current data collection paradigms, reliant on in-prompt pairwise annotations, suffer from labeling noise. Concurrently, the architectural design of VLM-based RMs, particularly their output mechanisms, remains underexplored. Furthermore, RM is susceptible to reward hacking in post-training. To mitigate these limitations, we propose {\ours}, a systematic framework for video RM training. Our framework first sources high-quality, cost-efficient data via single-item binary annotations, then constructs preference pairs using a cross-prompt pairing strategy. Architecturally, we employ a Hierarchical Progressive Query Attention mechanism to enhance feature aggregation. Finally, we introduce a modified BT loss that explicitly accommodates win-tie scenarios. This approach regularizes the RM's score distribution for positive samples, providing more nuanced preference signals to alleviate over-focus on a small number of top-scoring samples. Our approach is validated on benchmarks evaluating physical plausibility, subject deformity, and semantic alignment, demonstrating improvements in direct RM evaluation metrics and in the efficacy of post-training on video generation models. Code and benchmark are available at \url{https://github.com/lian700/SoliReward}
\end{abstract}

\section{Introduction}
\label{sec:intro}
Recent advancements in video generation models, such as Sora 2~\cite{openai_sora2}, Veo 3~\cite{google2025veo}, and Seedance~\cite{gao2025seedance}, have exhibited remarkable capabilities in synthesizing high-fidelity, temporally coherent visual content. Behind these achievements, post-training alignment techniques, analogous to Reinforcement Learning from Human Feedback (RLHF)~\cite{ouyang2022training} in Large Language Models (LLMs), are becoming a key driving force for enhancing model performance and correcting physical implausibility, visual artifacts and instruction following. Currently, methods such as DanceGRPO~\cite{xue2025dancegrpo} and LongChat-Video~\cite{team2025longcat} have begun to utilize flow-based GRPO~\cite{liu2025flow} methodologies to align video models.

However, the effectiveness of these alignment techniques relies heavily on the capabilities of their core component: the Reward Model (RM). RM is trained to quantify human preferences. However, constructing an RM that can accurately capture complex video qualities faces critical issues. These include data annotation noise and inconsistency, the vulnerability to reward hacking, and the underexplored design space for VLM-based RM architectures.
On the data front, the dominant paradigm, pairwise preference learning~\cite{liu2025improving}, trains on video pairs $(y_w, y_l)$ where $y_w$ is preferred over $y_l$, typically using the Bradley-Terry (BT) loss~\cite{bradley1952rank}. While effective, this ``in-prompt'' annotation (pairing videos from the same prompt) struggles with pairs of comparable quality, where annotator ambiguity introduces substantial label noise. 
An alternative, point-wise scoring (e.g., using a five-level scale)~\cite{liang2024rich}, avoids direct comparison but introduces high inter-annotator disagreement due to the subjective and ambiguous nature of intermediate scores. This noise in the preference data fundamentally degrades the RM's ability to accurately rank video quality. Beyond data noise, RM is highly susceptible to reward hacking in post-training, where the learned proxy objective deviates from intended human preferences~\cite{skalse2022defining}. Finally, the architecture used to extract this scalar reward is often insufficiently expressive, leading to a collapse of the reward where scores cluster together. Common methods, such as using the last token's embedding~\cite{zhao2025swift}, relying on the output of a dedicated special token~\cite{liu2025improving}, or using the probability of a ``yes/no" token~\cite{wu2025rewarddance}, may fail to capture the full spectrum of information, from low-level artifacts to high-level semantic alignment, encoded across the model's layers.

To mitigate these challenges, we propose a systematic framework towards robust video reward model training, including data annotation, training strategy and model architecture. 
We reorient data collection, shifting from complex relative comparisons or ambiguous multi-level scores towards simple, low-noise \textbf{single-item binary annotations} (Pass / Fail) targeting specific quality dimensions. To leverage the ranking ability of BT loss with these binary labels, we introduce a \textbf{cross-prompt pairing strategy}, where videos within one pair could belong to different prompts. This strategy generates a large-scale, high-signal preference dataset from simple binary labels, circumventing the noise inherent in ambiguous in-prompt comparisons.

Furthermore, we propose the BT with Win-Tie (BT-WT) loss to alleviate the critical issue of reward hacking~\cite{skalse2022defining}. By augmenting standard win-lose pairs with `win-tie' pairs (e.g., pairing two `Pass' samples), it explicitly penalizes score variance within the positive set, forcing the RM to map all high-quality samples to a compact manifold  and thus mitigating reward spikes that lead to hacking.

Finally, to address the architectural limitations identified in extracting a reward signal from a VLM~\cite{bai2025qwen2} backbone, we move beyond simplistic pooling strategies. We propose the \textbf{Hierarchical Progressive Query Attention (HPQA)}, a novel architecture that explicitly aggregates and refines features from multiple transformer layers, fusing low-level visual fidelity with high-level semantic understanding for a more robust reward signal. 
Our contributions are as follows:
\begin{itemize}
    \item \textbf{On the data level,} we design a highly efficient and low-noise pipeline combining single-item binary annotation with a novel cross-prompt pairing strategy.
    \item \textbf{For the training objective,} we introduce BT-WT loss to enhance data efficiency and demonstrably mitigate reward hacking by balancing positive sample distributions.
    \item \textbf{Architecturally,} we propose the \textbf{HPQA}, a new reward model architecture that fuses multi-layer features for a more robust scalar reward.
    \item Finally, we establish new benchmarks for evaluating video generation quality based on \textbf{subject deformity} and \textbf{physical plausibility}, and demonstrate through extensive experiments that our combined framework surpasses existing RM training and architectural baselines.
\end{itemize}

\begin{figure*}[!tb]
    \centering
    \includegraphics[width=1.0\linewidth]{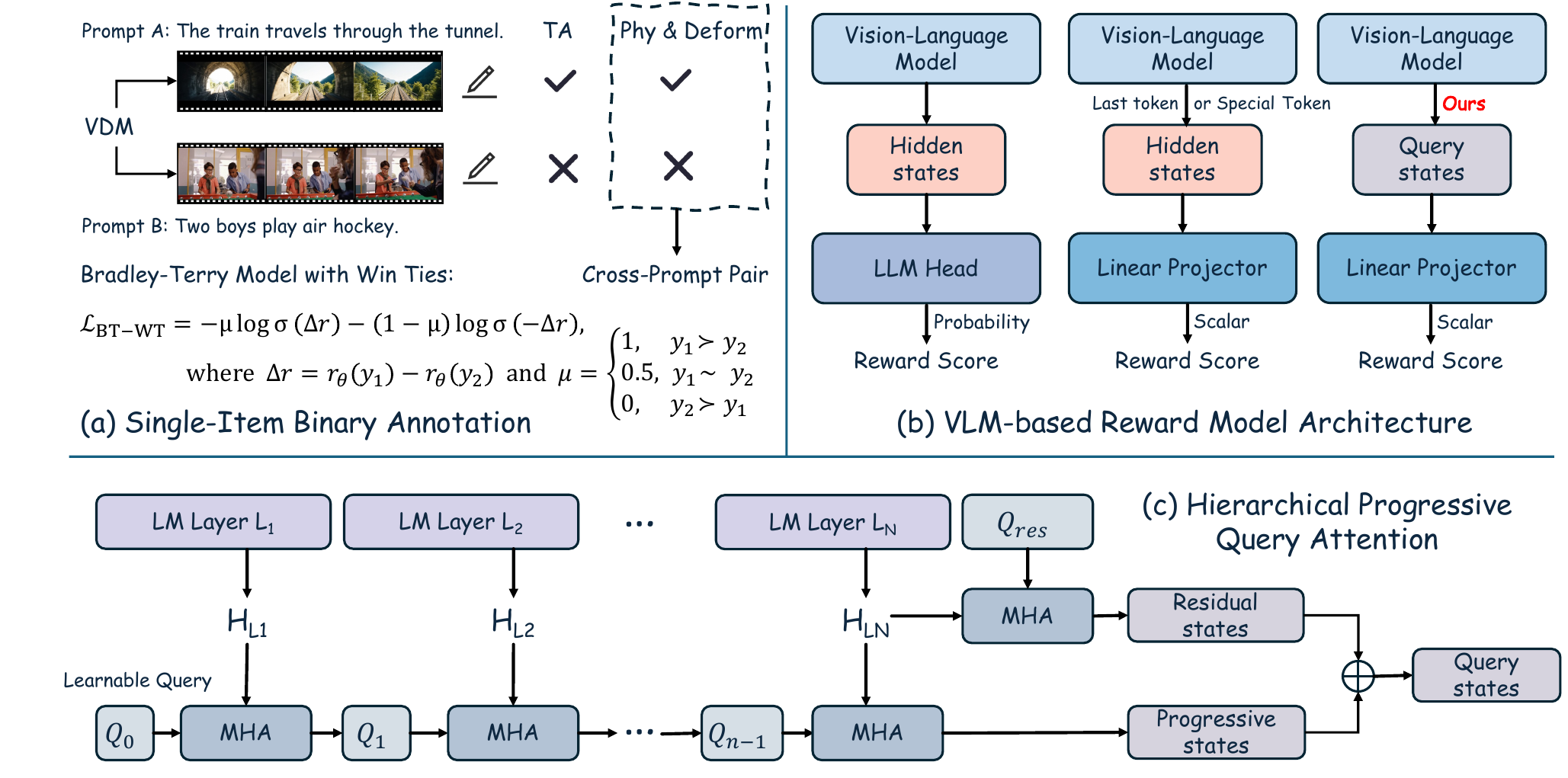}
    \caption{
    Pipeline of {\ours}, our framework for data annotation and training of video reward models.
(a) We introduce a single-item binary annotation method, coupled with a cross-prompt pairing strategy, to mitigate annotation noise. Furthermore, to alleviate reward hacking, we propose the Bradley-Terry with Win-Tie (BT-WT) loss.
(b/c) We propose a novel VLM-based Reward Model (VLM-RM) architecture, featuring a Hierarchical Progressive Query Attention (HPQA) adapter. This adapter progressively aggregates multi-level representations from the VLM backbone to compute a robust reward score.
    }
\vspace{-12pt}
    \label{fig:pipeline}
\end{figure*}

\section{Related Works}
\subsection{Data Construction for Reward Model Training}
The efficacy of a Reward Model is fundamentally constrained by its training data. Existing annotation paradigms generally fall into three categories: pairwise comparisons, point-wise scoring, and binary annotation.

\textbf{Pairwise Preference Annotation.} The dominant paradigm in RLHF~\cite{ouyang2022training} and recent video RMs~\cite{kirstain2023pick,ma2025hpsv3} involves querying preferences between two generations from the \textit{same} prompt. This naturally aligns with BT objectives~\cite{bradley1952rank}, which excel at modeling relative utility. However, in-prompt annotation suffers from severe decisional ambiguity when candidates exhibit similar quality levels. It forces subjective tie-breaking, injecting noticeable noise. While some efforts introduce ``tie" options~\cite{liu2025improving}, defining standardized thresholds for neutrality remains difficult and does not fully resolve the underlying ambiguity.

\textbf{Point-wise Scoring Annotation.}
Alternatively, discrete multi-level Likert scales (e.g., VideoScore~\cite{he2024videoscore}) could introduce high-variance labeling noise, particularly at subjective decision boundaries. As human preference is intrinsically continuous~\cite{you2025teaching}, forcing rigid discrete labels onto borderline samples (e.g., `3' vs. `4' for similar inputs) discards uncertainty and creates artificial quality gaps. This limitation is underscored by VideoScore~\cite{he2024videoscore}, which reports critically low Inter-Annotator Agreement~(IAA) (e.g., Fleiss' $\kappa < 0.1$~\cite{fleiss1973equivalence} in Trial 2). We attribute this poor agreement to these boundary inconsistencies.

\textbf{Binary Annotation.} Recent work such as VisionReward~\cite{xu2024visionreward} demonstrates that simplifying tasks to \textbf{binary} checklists achieves annotation consistency ($\sim89\%$ agreement). 
However, its BT-objective is confined to learning linear weights, degenerating the reward into discrete scores and lacking the ranking granularity of more general RMs~\cite{liu2025improving}.
Our work bridges this critical gap. 

\subsection{Architecture of Reward Model}
Vision Language Models (VLMs)~\cite{bai2025qwen2} have emerged as the primary architectural choice for reward models, such as ~\cite{ma2025hpsv3,liu2025improving,wu2025rewarddance,xu2024visionreward,he2024videoscore}.
Commonly, a Vision Transformer processes visual inputs, mapping them into the language embedding space. A language backbone then uniformly processes these visual representations alongside textual inputs.
However, the methodology for extracting a scalar reward score from the VLM's output representations remains an open question. Three primary approaches are commonly employed to derive this score:

\textbf{Linear Head}~\cite{zhao2025swift}. This method utilizes the embedding of the \textit{last token} from the VLM's final hidden state, which is then fed into a Multi-Layer Perceptron (MLP) to produce the scalar score.

\textbf{Special Token Head}~\cite{liu2025improving}. A dedicated special token (e.g., \texttt{<VQ>, <MQ>, <TA>} in VideoAlign) is prepended to the input sequence. The score is constructed by selectively isolating and extracting the logits from these pre-defined special token positions, with each token's output mapping directly to a specific reward dimension.

\textbf{Token Probability}~\cite{wu2025rewarddance}. This method reframes the task as a prompted binary decision. The VLM is instructed to evaluate whether the visual input meets specific criteria and to respond with a binary answer (e.g., ``yes'' or ``no''). The reward is then extracted from the probability/logits assigned to the ``yes'' token in the model's output.

\subsection{Post-Training on Visual Generation}
Aligning generative models with human preferences is a considerable challenge. The standard paradigm, RLHF, first trains a RM on human preferences and then fine-tunes the generative policy to maximize the RM's score. 

\textbf{Group Relative Policy Optimization.}
Adapted for visual generation, GRPO-based online algorithms~\cite{xue2025dancegrpo,liu2025flow,li2025branchgrpo,luo2025sample,zhong2026euphonium} addresses the instability of standard policy gradient methods. They utilize online feedback from reward model as guidance for optimization. By introducing Brownian motion, the ordinary differential equation (ODE) inherent in flow matching~\cite{lipman2022flow} is transformed into a stochastic differential equation (SDE). This infusion of stochasticity enhances the model's exploration within the generation space.

\textbf{Direct Preference Optimization.}
DPO~\cite{liu2025videodpo,liu2025improving,wu2025densedpo,wallace2024diffusion} offers an alternative to computationally intensive policy gradient methods by recasting the constrained RL objective as a classification problem. This reformulation simplifies the process, yielding a more stable training objective. RM can be leveraged to construct preference pairs for DPO training.

\textbf{Reward Feedback Learning.}
ReFL~\cite{xu2023imagereward,clark2023directly,shen2025directly,mi2025video} is an alignment technique that utilizes a differentiable reward model, employing the reward signal directly as the loss function. This mechanism allows for end-to-end optimization by backpropagating gradients from the reward model directly through the generative model's parameters, thus avoiding the complexities of GRPO.

\section{Methodology}
In this section, we detail our pipeline. As shown in \cref{fig:pipeline}, we first introduce our data annotation approach (\cref{sec:data-annotation}) and cross-prompt pairing strategy (\cref{sec:cross-prompt-pairing-strategy}). We then describe the training loss, a win-tie variant of the Bradley-Terry model (\cref{sec:bt-with-win-tie}) designed to mitigate reward hacking and the reward model architecture HPQA (\cref{sec:hierarchical-progressive-query-attention}).

\subsection{Data Annotation}
\label{sec:data-annotation}

The efficacy of an RM depends heavily on preference data quality, yet standard pairwise and point-wise methods suffer from high annotator ambiguity and disagreement~\cite{he2024videoscore,liu2025improving}. To mitigate this, we design a single-item binary annotation. By evaluating single videos against objective criteria (Pass/Fail), we diminish label noise and improve IAA, as shown in \cref{sec:exp-data-annotation}.
We focus on three critical dimensions: \textbf{Physical Plausibility}, \textbf{Subject Deformity}, and \textbf{Semantic Alignment}. This binary scheme yields high-confidence signals. We annotate $250\mathrm{k}$ in-house training videos and a $50\mathrm{k}$-video out-of-distribution (OOD) test set generated by other SOTA models. These samples originate from $20\mathrm{k}$ unique prompts (please refer to  Appendix~\cref{appendix:sec_annotation} for details).

\subsection{Cross-Prompt Pairing Strategy}
\label{sec:cross-prompt-pairing-strategy}
Based on the high-quality binary labels from \cref{sec:data-annotation}, we construct our preference data. We treat all samples within the ``Pass" ($W$) set as mutually equivalent in preference ($\forall y_i, y_j \in W, y_i \sim y_j$). While ``Fail" ($L$) samples may exhibit varying degrees of quality, every ``Pass" sample is definitively preferred over every ``Fail" sample ($\forall y_i \in W, \forall y_j \in L, y_i \succ y_j$). This establishes a clear preference order between the two sets.

Crucially, this formulation allows us to implement a cross-prompt pairing strategy. The Bradley-Terry model, upon which our loss is based, does not theoretically require pairs to originate from the same prompt~\cite{sun2024rethinking}. By pairing ``Pass" and ``Fail" samples from different prompts, we create a large-scale, diverse dataset that forces the RM to learn generalized representations of quality rather than simple in-prompt relative rankings.

\subsection{Bradley-Terry with Win-Tie}
\label{sec:bt-with-win-tie}
Reward model training relies on pairwise preference data. Given a pair $(y_i, y_j)$ where $y_i$ is preferred over $y_j$, the RM is commonly trained via the Bradley-Terry (BT) loss:
\begin{equation}
    \mathcal{L}_{\mathrm{BT}} =  \mathbb{E}_{(y_i, y_j) \in D} \left [ -\log(\sigma(r_\theta(y_i) - r_\theta(y_j))) \right]
    \label{eq:bt_loss}
\end{equation}
However, during data annotation, annotators may deem two samples to have near-identical quality (i.e., ties, $y_i \sim y_j$). The vanilla BT loss discards these tie samples, reducing data efficiency. Furthermore, this win-lose only training scheme, particularly when annotating on specific dimensions, can induce reward hacking during subsequent model post-training~\cite{skalse2022defining}. Training exclusively on win-lose pairs seeks to maximize the reward margin $r_\theta(y_i) - r_\theta(y_j)$ in \cref{eq:bt_loss}. Crucially, this objective imposes no constraints on the reward distribution \emph{within} the set of positive samples.

This absence of constraint permits the RM to assign disproportionately high rewards to certain positive samples exhibiting shortcut features, while assigning relatively lower rewards to other valid positive samples that lack these artifacts. When a subsequent generative model explores this reward landscape, it preferentially converges to these reward spikes, learning to generate hacking-type samples. This exploitation of the RM's scoring deficiencies to achieve high rewards, despite producing outputs that misalign with true human preferences, constitutes reward hacking.




Therefore, to mitigate reward hacking and improve data efficiency, we propose the Bradley-Terry with Win-Tie (BT-WT) strategy. We supplement the win-lose pairs ($y_i \succ y_j$) with win-tie pairs ($y_i \sim y_j$), which are constructed by pairing positive samples with other positive samples. Corresponding loss function is modified as follows, where $W$ and $L$ are the sets of positive and negative samples, respectively:
\begin{equation}
\label{eq:bt_wt_loss}
\begin{aligned}
    \mathcal{L}_{\mathrm{BT-WT}} &= \mathbb{E}_{(y_i, y_j) 
    \in W \times (W \cup L)} \left[ - \mu \log \sigma(\Delta r) \right. \\
    & \left.  - (1 - \mu) \log \sigma(-\Delta r) \right] \\
    \Delta r &= r_\theta(y_i) - r_\theta(y_j),
    \mu = 
    \begin{cases}
        1,   & y_i \succ y_j \\
        0.5, & y_i \sim y_j
    \end{cases}
\end{aligned}
\end{equation}

The fundamental mechanism by which these win-tie pairs mitigate reward hacking is the imposition of a regularization constraint on the RM's output space. As shown in \cref{eq:bt_wt_loss}, the tie-loss component of $\mathcal{L}_{\mathrm{BT-WT}}$ explicitly penalizes the reward discrepancy between any two positive samples.
This loss term compels the RM to map all positive samples onto a \textbf{more compact and dense manifold} within the reward space, ensuring $r_\theta(y_i) \approx r_\theta(y_j)$. This mechanism acts as a regularizer, effectively flattening the spurious spikes in the reward landscape and promoting smoothness.

\begin{figure}[!tb]
    \centering
    \vspace{-12pt}
    \includegraphics[width=1.0\linewidth]{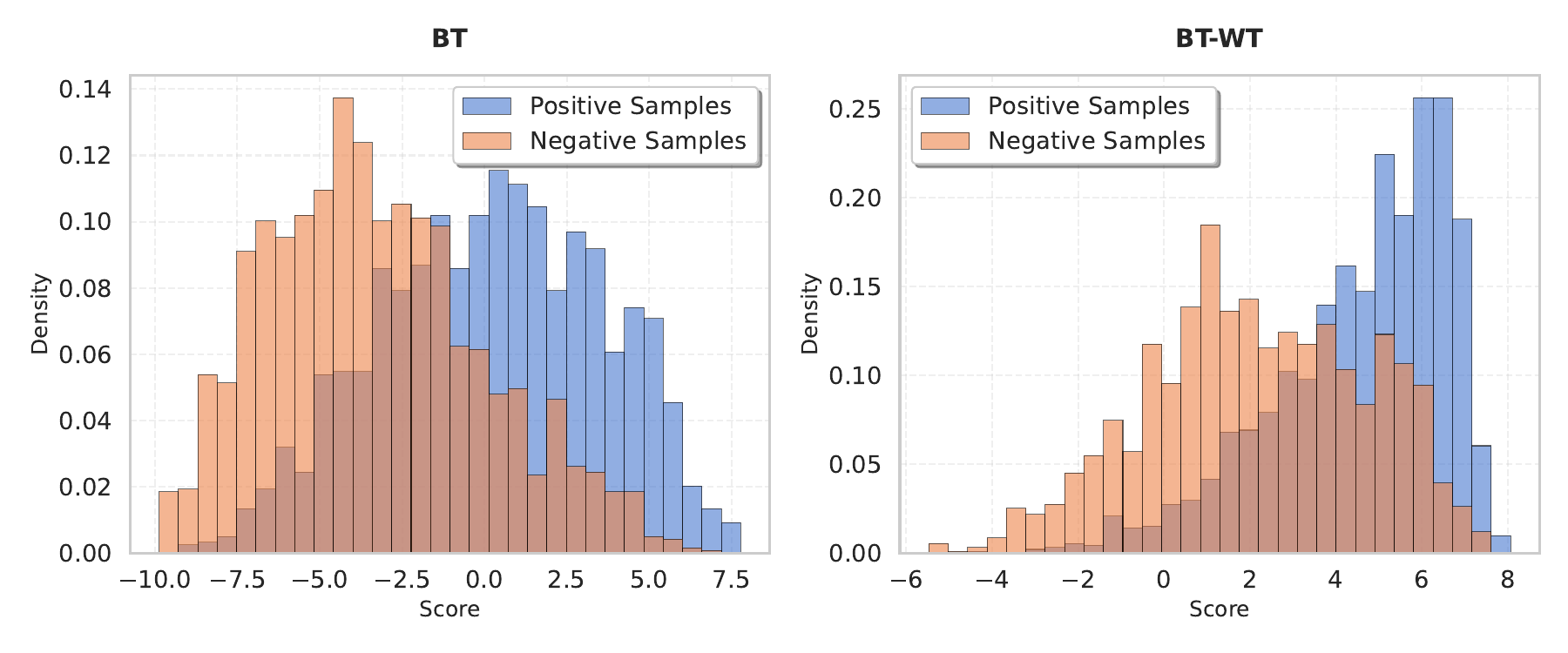}
    \caption{Reward distribution for BT and BT-WT. BT-WT contributes to a more concentrated distribution in the high-score segment for positive samples.}
    \label{fig:score_dist_bt_vs_btwt}
\end{figure}

Consequently, RM provides a more robust and accurate optimization signal for the generative model. This guides the model to learn the generalizable features of high-quality samples, rather than exploiting the specific defects and loopholes of the reward model.
As shown in ~\cref{fig:score_dist_bt_vs_btwt}, incorporation of win-tie sample pairs markedly alters the score distribution of positive samples, leading to a more concentrated distribution in the high-score segment. We also propose a test to evaluate a dimension's suitability for win-tie pair construction, detailed in \cref{sec:whether-pair-win-tie} in Appendix.

In contrast, VideoAlign~\cite{liu2025improving} retains all tie samples and employs a loss function that operates on A-wins, B-wins, or a tie (including both win-tie and lose-tie pairs). We argue this strategy is unsuitable for our annotation scenario. Given two independently labeled negative samples (e.g., for deformity), we cannot assert their degree of deformity is equivalent. Erroneously pairing them as a tie would diminish the discriminative capacity of RM and degrade model performance.

\subsection{Hierarchical Progressive Query Attention}
\label{sec:hierarchical-progressive-query-attention}
We introduce Hierarchical Progressive Query Attention (HPQA), a novel reward model adapter architecture for computing the reward scalar from query states. HPQA computes rewards by aggregating features from multiple language model (LM) transformer layers. Let $H_i \in \mathbb{R}^{B \times S \times D}$ be the hidden state output of layer $i$, and $I = [l_1, l_2, \dots, l_N]$ be a list of $N$ specified layer indices.

The operational flow begins by generating an initial query vector $q^{(1)} \in \mathbb{R}^{B \times 1 \times D}$. This is achieved by attending to the first specified hidden state $H_{l_1}$ using a single, learnable query vector $q^{(0)} \in \mathbb{R}^{1 \times 1 \times D}$. This vector $q^{(0)}$ acts as the Query, while the input hidden state $H_{l_1}$ serves as both the Key and Value in a standard Multi-Head Attention (MHA) operation:
\begin{equation}
q^{(1)} = \mathrm{MHA}_{1}(Q=q^{(0)}, K=H_{l_1}, V=H_{l_1}) 
\label{eq:init_query}
\end{equation}
The output $q^{(1)}$ is progressively refined over the subsequent $N-1$ layers specified in $I$. For $i=2$ to $N$, the MHA module at layer $l_i$ takes $q^{(i-1)}$ as Query, using hidden state $H_{l_i}$ as both Key and Value:
\begin{equation}
q^{(i)} = \mathrm{MHA}_{i}(Q=q^{(i-1)}, K=H_{l_i}, V=H_{l_i})
\label{eq:progressive_refinement}
\end{equation}
The final iteration's resulting vector $q^{(N)}$ serves as the progressive feature $q_{\mathrm{prog}}$. Concurrently, $o_{\mathrm{res}}$ is computed by attending to the last hidden state $H_L$ via a separate attention module. This module uses another learnable query $q_{\mathrm{res}}$ as the Query, with $H_L$ serving as both Key and Value:
\begin{equation}
o_{\mathrm{res}} = \mathrm{MHA}_{\mathrm{res}}(Q=q_{\mathrm{res}}, K=H_L, V=H_L)
\label{eq:final_query}
\end{equation}
These two vectors are combined via residual connection and passed to $\mathrm{RewardHead}$ to yield the final scalar reward $r$:
\begin{equation}
r = \mathrm{RewardHead}(q_{\mathrm{prog}} + o_{\mathrm{res}})
\label{eq:final_reward}
\end{equation}

A primary advantage of this mechanism is its ability to create a rich feature representation by explicitly aggregating information hierarchically. This approach is conceptually grounded in findings from linguistic analysis in LLMs~\cite{vig2019analyzing}, which reveals that transformer layers exhibit functional specialization. For instance, their analysis indicates that attention aligns most strongly with syntactic dependency relations in the middle layers , while the deepest layers are most effective at capturing distant relationships. Our mechanism is designed to explicitly harness this hierarchical specialization. The progressive refinement \cref{eq:progressive_refinement} allows the model to synthesize a query that explicitly bridges these different semantic levels, fusing low-level fidelity with high-level abstraction. Furthermore, the residual connection ensures that this hierarchical information augments rather than replaces the representation from the final layer. 


\input{table/IAA}

\section{Experiments}
In this section, we empirically evaluate our reward model on both in-domain and out-of-domain accuracy. Additionally, evaluations are demonstrated in post-training on text-to-video (T2V) task. We use the InternVL3~\cite{zhu2025internvl3} series as the reward model backbone and HunyuanVideo~\cite{kong2024hunyuanvideo} for post-training verification.

\subsection{Data Annotation and Pairing Strategy}
\label{sec:exp-data-annotation}
To validate the reliability of our single-item binary annotation approach against traditional pairwise comparisons, we conduct a formal Inter-Annotator Agreement (IAA) analysis, tasking 5 annotators to label a subset of data using both methods.
In~\cref{tab:iaa_results}, we report Krippendorff's $\alpha$~\cite{Hayes01042007}, Fleiss's $\kappa$~\cite{fleiss1973equivalence}, and the raw percentage agreement. Our single-item task achieves a \textbf{Moderate} agreement, registering $\alpha = 0.4939$, $\kappa = 0.4925$, and a strong raw agreement of 77.33\%. These results are substantially higher than those from the pairwise comparison task, which only reached \textbf{Fair} agreement ($\alpha = 0.3516$, $\kappa = 0.3494$) with a 54.67\% raw agreement. It confirms that our point-wise annotation framework yields more consistent and reliable labels from annotators compared to the conventional pairwise approach.

Regarding the pairing strategy, a comparison between in-prompt and cross-prompt methods is deferred to \cref{sec:pairing-strategy} in Appendix (\cref{tab:RM-pair_strategy-ACC-ID-OOD} and \cref{tab:RM-pair_strategy-margin-ID-OOD}) due to page limitations. The cross-prompt strategy achieves performance comparable to its in-prompt counterpart, while also effectively leveraging data from prompts that yielded only a single video. This approach enhances data utilization without compromising reward model performance.

\subsection{Reward Model}
\subsubsection{Evaluation of Accuracy}
\input{table/RM-compared_to_baseline}

\begin{figure}[!tb]
    \centering
    \includegraphics[width=1.0\linewidth]{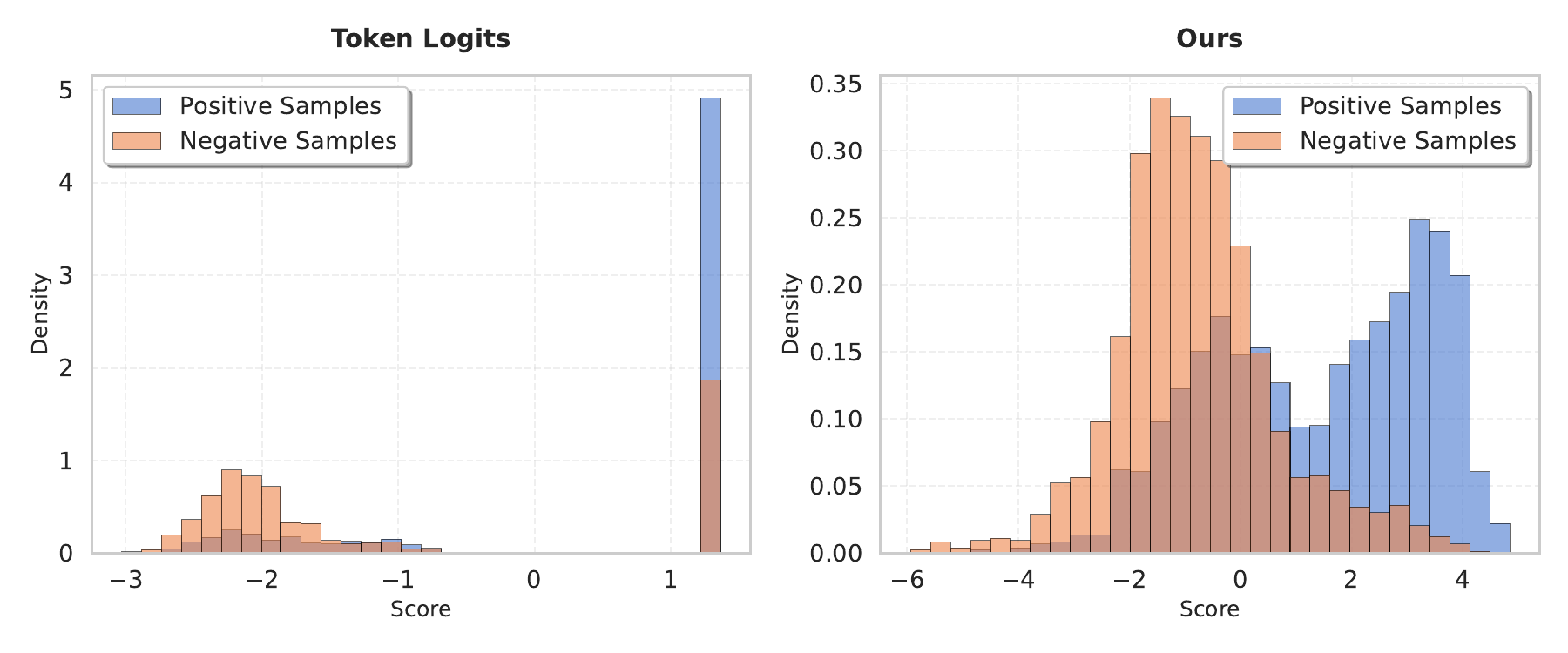}
    \caption{
    The reward score distributions on the semantic alignment task reveal that alternative architectures suffer from severe score clustering, assigning identical ratings to many samples.
    }
    \label{fig:score_dist_architecture}
\end{figure}

We benchmark the accuracy of {\ours} against representative baselines on both in-domain (ID) and out-of-domain (OOD) data. ID accuracy is measured on a held-out partition of our training dataset, which is unseen during training process. For OOD evaluation, we employ videos generated by state-of-the-art (SOTA) models (including Wan2.1~\cite{wan2025wan}, Wan2.2~\cite{wan2025wan}, Veo 3~\cite{google2025veo} and Seedance 1.0~\cite{gao2025seedance}) from a curated prompt set. Video quality is then assessed by human annotators across three criteria: Physical Plausibility, Subject Deformity (Phy \& Deform), and Semantic Alignment (TA).
Evaluation on OOD dataset is critical~\cite{wu2025rewarddance} because ID accuracy is an insufficient indicator of downstream task performance. OOD accuracy, which assesses generalization capability on shifted distributions, serves as a more robust predictor of a model's final efficacy. Consequently, OOD accuracy emerges as a more critical metric for evaluating a model's true utility.

We present the reward model accuracy comparison in ~\cref{tab:RM-compared_to_baseline}. Our model demonstrates substantially superior performance and generalization capability compared to existing baselines. For the \textbf{Phy \& Deform} task, our model achieves the highest accuracy on both ID data (78.48) and OOD data (80.08). This surpasses all baselines, including the second-best model, VideoAlign (71.60). On the \textbf{TA} task, our model attains a dominant ID accuracy of 79.02, far exceeding the closest baseline, VideoPhy (54.85). Its OOD performance (60.25) is highly competitive, matching the top (VideoPhy, 60.52) and outperforming others. 

Notably, several baselines (marked with $^*$) that suffer from score clustering (e.g., LiFT, UnifiedReward). They output discrete scores, such as integers from 1-5 or categorical labels (good/normal/bad). This coarse granularity, coupled with their limited OOD generalization, causes score clustering: many samples are assigned identical ratings, thus deflating the accuracy.

\subsubsection{Bradley-Terry with Win-Tie}
\label{sec:exp-bt_wt}

Despite the reward models trained via BT and BT-WT exhibiting comparable accuracy, the BT-WT RM yields better performance during post-training (shown in \cref{tab:bt_vs_btwt}). We analyze this divergence from the following two perspectives.

\input{table/RM-BT_BTWT}

\begin{figure}[!tb]
    \centering
    \vspace{-4pt}
    \includegraphics[width=0.8\linewidth]{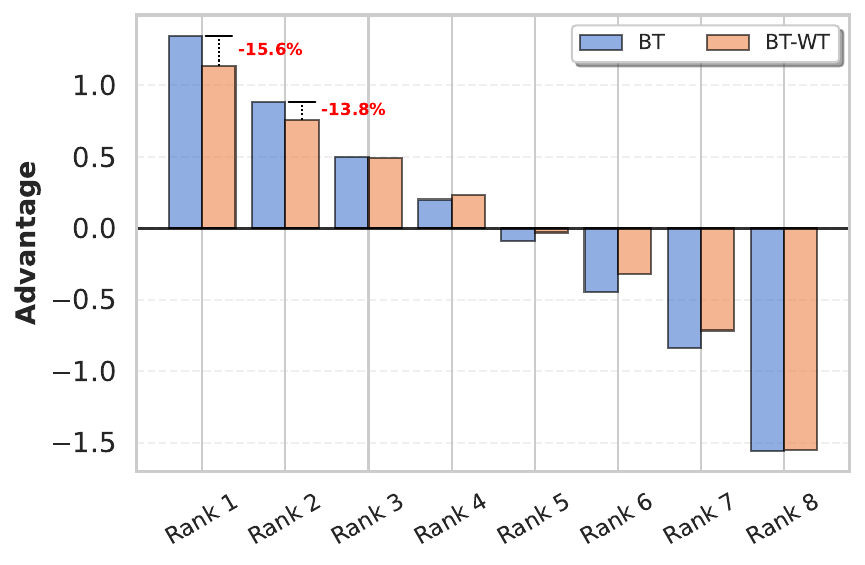}
    \caption{Intra-group advantage distribution in BT-WT exhibits smaller advantages for top-ranked samples compared to BT, thereby mitigating the over-optimization.}
    \label{fig:advantage_distribution_comparison}
\end{figure}

\textbf{Group Advantage.} Our empirical observations confirm that group advantage from a standard BT-trained reward model exhibit larger absolute values than those from BT-WT-trained RM, especially in top-rank samples. Standard BT loss, seeking only to maximize the positive-negative margin, induces a sharp, high-contrast scoring reward margin within the high-quality samples. Consequently, during GRPO-based post-training leveraging group advantage, this RM assigns high-variance scores to samples of comparable quality within a group. This high variance in advantage calculation, $A_i = \frac{r_i - \bar{r}}{\sigma}$, leads to \textbf{over-optimization}: outlier samples with anomalously high scores generate massive positive advantages, causing the policy to overfit to non-generalizable features. The win-tie pairs in BT-WT act as a calibrator, compelling the RM to map high-quality samples to a more compact score interval. This results in smaller absolute advantage values, which precisely signifies lower gradient variance for more robust policy updates. Group advantage comparison between BT and BT-WT during post-training is shown in \cref{fig:advantage_distribution_comparison}.

\textbf{Regularization.} From the perspective of regularization, the win-tie pairs function as a critical regularizer against reward hacking. An RM trained exclusively on win-lose pairs is prone to overfitting to shortcut features. This vulnerability is readily exploited during post-training: the policy network quickly identifies and leverages these spurious features to maximize corresponding reward, while achieving only limited improvements in perceptual video quality. The BT-WT paradigm, by introducing win-tie pairs, compels the RM to learn a more robust and generalizable internal representation aligned with human preferences. This paradigm constrains the RM to assign proximate scores to perceptually distinct yet high-quality samples. This constraint enhances the RM's capacity to model the complex concept of human preference, making it less susceptible to reward hacking and yielding a more reliable optimization signal for post-training.

\input{table/post_training_performance}
\input{table/RM-architecture}

\subsection{Post-Training}
We conduct post-training to further validate our reward model on physical plausibility and subject deformity. HunyuanVideo~\cite{kong2024hunyuanvideo} is selected as the video generation backbone. 
We apply online reinforcement learning post-training with DanceGRPO~\cite{xue2025dancegrpo}, guided by our reward model. This process enhances video quality in these two aspects. To quantify this, we employ the VideoAlign Motion Quality (MQ)~\cite{liu2025improving} and VBench2 Human Fidelity~\cite{zheng2025vbench} evaluation metrics. As shown in ~\cref{tab:post_training_performance}, the results confirm the efficacy of our reward model during this post-training phase.

\begin{figure*}[!tb]
    \centering
    \includegraphics[width=1.0\linewidth]{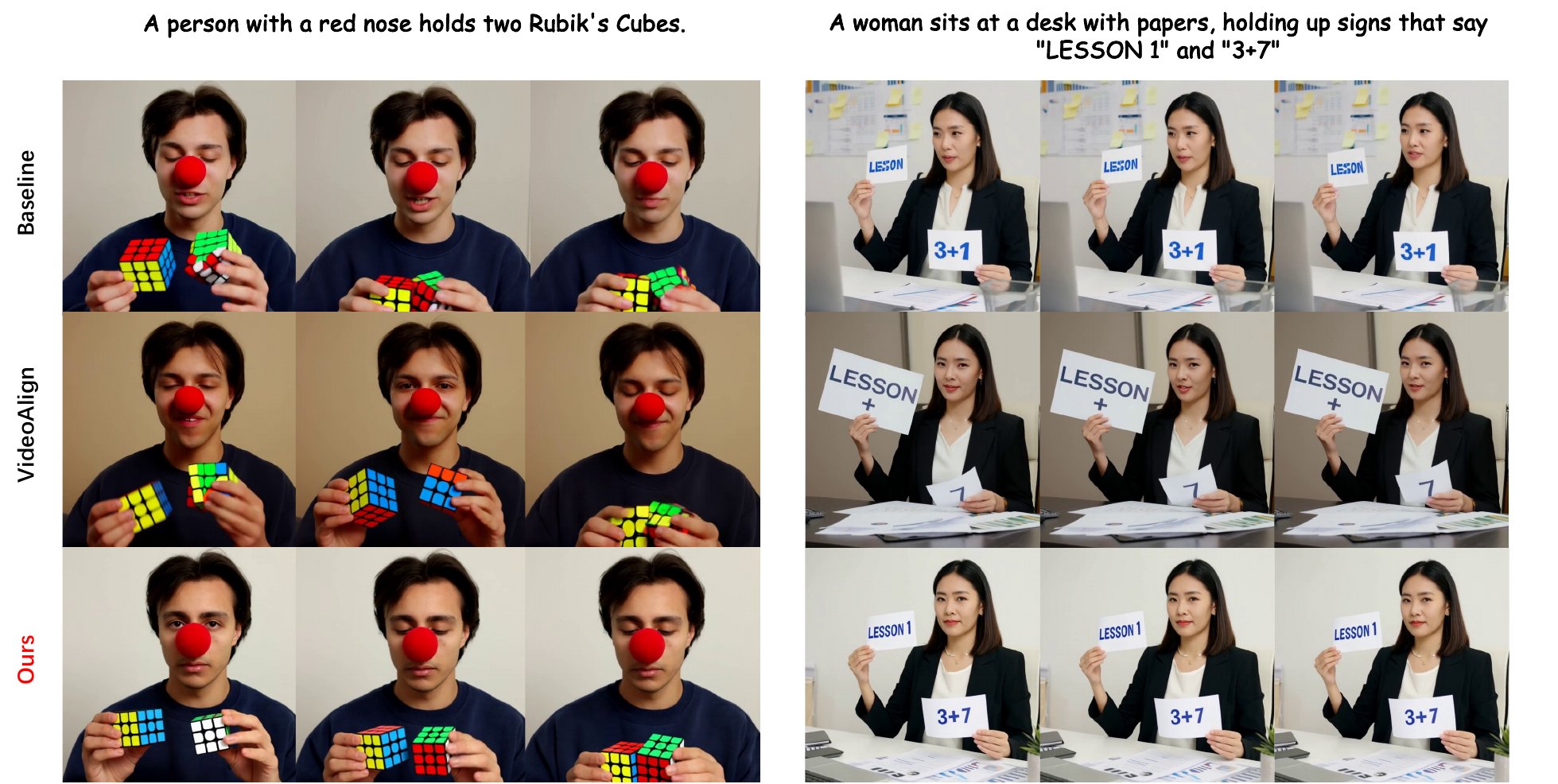}
    \caption{Visual results guided by different reward models. From top to bottom: Baseline (HunyuanVideo), VideoAlign, and Ours.}
    \vspace{-12pt}
    \label{fig:phy_deformity_case1}
\end{figure*}

\begin{figure}[!tb]
    \centering
    \includegraphics[width=1.0\linewidth]{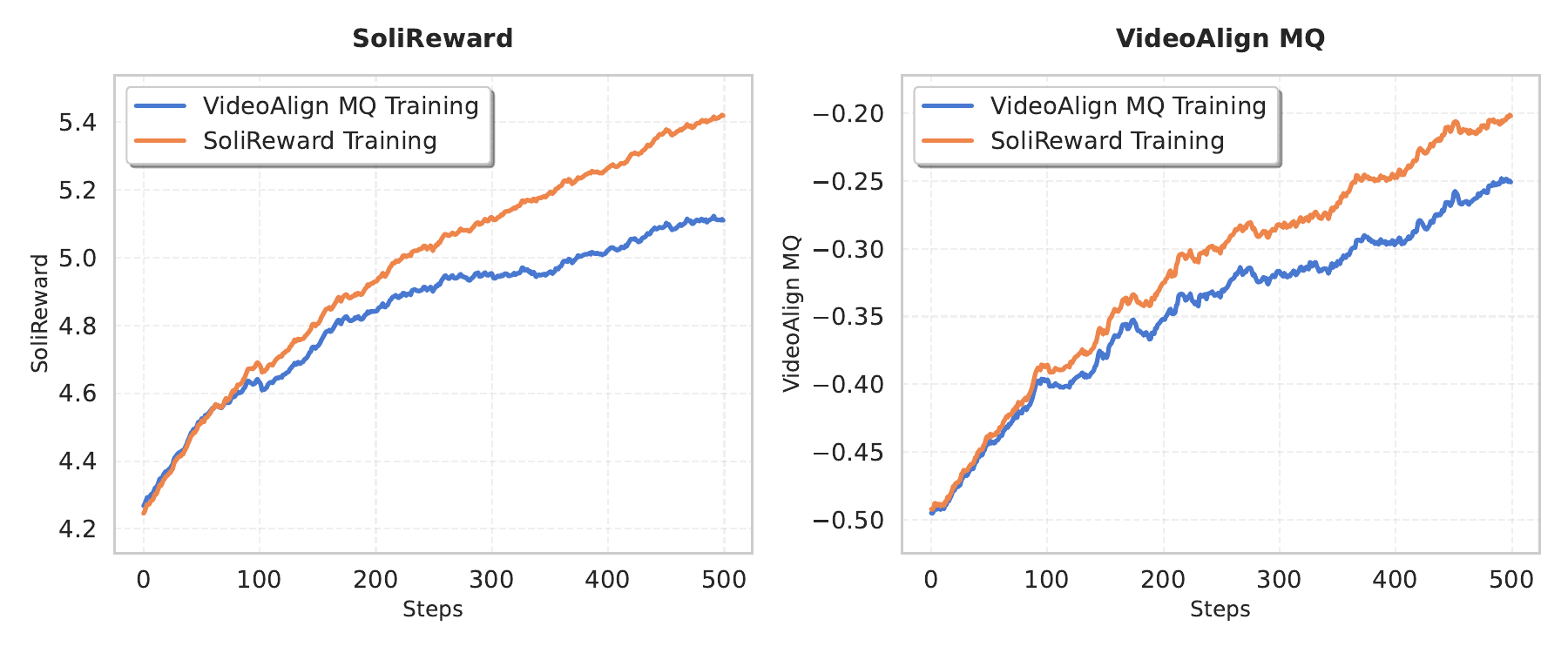}
    \caption{Comparison of reward scores between post-training guided by our reward model and VideoAlign MQ.}
    \label{fig:training_curve_MQ_vs_TRL}
\end{figure}

To further demonstrate the necessity of our reward model, we conduct a comparative analysis against existing RMs, such as VideoAlign MQ. We perform a parallel post-training experiment using VideoAlign MQ as the sole guidance signal. 
As shown in \cref{tab:post_training_performance} and \cref{fig:training_curve_MQ_vs_TRL}, VideoAlign MQ fails to effectively mitigate videos exhibiting physical law violations or deformities. Visual comparisons confirm this: models optimized with VideoAlign MQ continue to generate such artifacts. In contrast, under identical settings, the model guided by our reward signal successfully alleviates these issues in \cref{fig:phy_deformity_case1}. Furthermore, when evaluating the post-trained models using our reward model, VideoAlign MQ, and VBench2 Human Fidelity as distinct metrics, our approach consistently outperforms the VideoAlign-guided baseline across all three dimensions. This substantiates our claim that existing MQ-focused reward models are insufficient for addressing physical plausibility and subject deformity, a critical gap our reward model successfully fills.
Due to page limitation, post-training experiments on semantic alignment are demonstrated in ~\cref{sec:post_training_TA} in Appendix.

\subsection{Ablation Study}

\textbf{Model Architecture.}
We benchmark our HPQA against three common alternative architectures: 1) Extracting the last token embedding from the final hidden state and projecting it via a linear layer to derive a score; 2) VideoAlign~\cite{liu2025improving}, which introduces a special token placeholder in the model input and uses its corresponding output embedding, processed by a linear layer, as the score; and 3) RewardDance~\cite{wu2025rewarddance}, which instructs the reward model to output `yes' or `no' and utilizes the predicted probability of the `yes' token as the reward. As ~\cref{tab:RM-architecture} indicates, our HPQA method outperforms the other architectures in terms of accuracy, validating its efficacy.

More critically, we observe that reward models based on these alternative architectures are highly susceptible to score clustering during training, particularly for semantic alignment. Specifically, the scores for positive and negative samples converge to several discrete values. This leads to an overly concentrated score distribution lacking sufficient diversity, consequently diminishing the model's ability to discriminate between samples of varying quality. This ultimately impairs its guidance capabilities during post-training optimization. In contrast, our HPQA architecture does not suffer from this instability, remaining stable across diverse video quality evaluation dimensions. This demonstrates the robustness of our proposed architecture.

\input{table/RM-BT-BCE}
\textbf{Binary Cross Entropy.}
RewardDance~\cite{wu2025rewarddance} augments the BT loss with a cross-entropy penalty during point-wise RM training. However, it does not thoroughly explore this component's impact on RM accuracy and subsequent post-training performance. Motivated by this gap, we empirically investigate the effect of this binary cross-entropy (BCE) penalty, treating the RM's output as logits for a binary pass/fail task. The results reveal a critical insight. RM accuracy is a deceptive metric when decoupled from the reward margin. As shown in \cref{tab:bce_results}, adding the BCE penalty (`BT-WT + BCE') yields comparable accuracy to our `BT-WT' baseline. However, this BCE-penalized RM leads to \textit{inferior} post-training performance.

We hypothesize this failure stems from \textbf{reward margin degradation}. The BCE loss compels the RM to function as a binary classifier, collapsing the nuanced score spectrum into two discrete points (e.g., positive vs. negative). This classification focus overshadows the fine-grained ranking needed to distinguish between samples, compromising the model's ability to express relative preference. This hypothesis is strongly supported by ~\cref{tab:bce_results}. The `BT-WT + BCE' model's reward margin (2.97) represents a 19.62\% collapse compared to the standard `BT-WT' model (3.72). Furthermore, a pure `BCE' model yields the lowest margin of all. During post-training, a policy network generates videos with subtle quality differences. An RM with a collapsed margin is blind to these nuances and thus incapable of providing the effective gradient signals needed for successful optimization. This finding underscores that a large, discriminative reward margin, not just classification accuracy, is a critical prerequisite for a successful reward model.

\section{Conclusion and Future Work}
\label{sec:conclusion}
We introduce {\ours}, a systematic framework for video reward model, including both data annotation and model training designed to mitigate annotation noise and reward hacking. It employs a scalable, low-noise, and cost-efficient single-item binary annotation scheme, enabling a robust cross-prompt pairing strategy that compels the RM to learn generalizable video quality representations. Technically, {\ours} features two key innovations: the Hierarchical Progressive Query Attention architecture, a reward adapter that fuses multi-level features, and the Bradley-Terry with Win-Tie loss to alleviate reward hacking. Experiments demonstrate {\ours} achieves superior RM accuracy and, crucially, translates these gains to improved post-training performance.


\textbf{Limitation \& Future Work.}
Future work can be extended to other conditional generation tasks, such as image-to-video (I2V). 
Besides, our reward model scores input videos along a single dimension, and multiple dimensions can be fused into a single RM by designing several learnable queries.
We observe more performance gains when scaling our RM from 1B to 8B parameters, but diminishing returns when scaling further to 14B. This finding on model scaling highlights a key avenue for future exploration.

{
    \small
    \bibliographystyle{ieeenat_fullname}
    \bibliography{main}
}

\input{appendix}
\end{document}

%% file: preamble.tex
\newcommand{\emphcell}[1]{\cellcolor{gray!20} \textbf{#1}}
\usepackage{enumitem}
\usepackage{tabularx}
\usepackage{booktabs}
\usepackage{array}  
\usepackage{adjustbox}
\usepackage{amssymb}
\usepackage{bbm}
\usepackage{multirow}
\usepackage{bm}
\usepackage{colortbl}
\usepackage{tabularx}
\usepackage[accsupp]{axessibility}



%% file: table/IAA.tex
\begin{table}[!tb]
\centering
\caption{
Inter-Annotator Agreement (IAA) analysis across 5 annotators. We report Krippendorff's $\alpha$~\cite{Hayes01042007}, Fleiss's $\kappa$~\cite{fleiss1973equivalence} and raw agreement for our binary \textit{single-item} task and \textit{pair-wise comparison} task. Single-item is more consistent than pairwise annotation.
}
\label{tab:iaa_results}
\adjustbox{width=0.9\linewidth}{
\begin{tabular}{l|ccccccccccc}
\toprule
\textbf{Annotation} & \bm{$\alpha$} & \bm{$\kappa$} & \textbf{Agreement} & \textbf{Interpretation} \\
\midrule
\emphcell{Single-Item} & \emphcell{0.4939} & \emphcell{0.4925} & \emphcell{0.7733} & \emphcell{Moderate} \\
Comparison         & 0.3516 & 0.3494 & 0.5467 & Fair \\
\bottomrule
\end{tabular}
}
\end{table}

%% file: table/RM-compared_to_baseline.tex
\begin{table}[!tb]
 \centering
 \caption{Reward model accuracy compared to baselines. $^*$ means the score distribution degenerates to discrete values.}
 \adjustbox{width=0.9\linewidth}{
 \begin{tabular}{ll|cc}
  \toprule
  \multirow{2}{*}{\textbf{Task}} & \multirow{2}{*}{\textbf{Approach}} & \multicolumn{2}{c}{\textbf{RM ACC}}\\
  \cmidrule(lr){3-4}
  ~ & ~ & \textbf{ID} & \textbf{OOD} \\
  \midrule
  \multirow{8}{*}{Phy \& Deform} & VideoScore~\cite{he2024videoscore} & 62.30 & 67.09 \\
  ~ & VideoScore-v1.1~\cite{he2024videoscore} & 60.20 & 68.65 \\
  ~ & LiFT~\cite{wang2024lift} & 52.60$^*$ & 61.11$^*$ \\
  ~ & VisionReward~\cite{xu2024visionreward} & 58.45 & 50.75 \\
  ~ & VideoPhy~\cite{bansal2024videophy} & 67.35 & 65.10 \\
  ~ & UnifiedReward~\cite{wang2025unified} & 49.90$^*$ & 44.18$^*$ \\
  ~ & VideoAlign~\cite{liu2025improving} & 54.40 & 71.60 \\
  ~ & \emphcell{Ours} & \emphcell{78.48} & \emphcell{80.08} \\
  \midrule
  \multirow{8}{*}{TA} & VideoScore~\cite{he2024videoscore} & 52.75 & 54.05 \\
  ~ & VideoScore-v1.1~\cite{he2024videoscore} & 51.20 & 53.36 \\
  ~ & LiFT~\cite{wang2024lift} & 1.75$^*$ & 7.40$^*$ \\
  ~ & VisionReward~\cite{xu2024visionreward} & 30.75$^*$ & 22.24$^*$ \\
  ~ & VideoPhy~\cite{bansal2024videophy} & 54.85 & \emphcell{60.52} \\
  ~ & UnifiedReward~\cite{wang2025unified} & 16.45$^*$ & 10.63$^*$ \\
  ~ & 
  VideoAlign~\cite{liu2025improving} & 49.50 & 49.14 \\
  ~ & \emphcell{Ours} & \emphcell{79.02} & {60.25} \\
  \bottomrule
 \end{tabular}
 }
 \label{tab:RM-compared_to_baseline}
\end{table}

%% file: table/RM-BT_BTWT.tex
\begin{table}[!tb]
\centering
\caption{Comparison of reward model performance trained via BT and BT-WT. Reward model accuracy and post-training evaluation metrics are reported.
}
\adjustbox{width=0.7\linewidth}{
\begin{tabular}{l|c|cccc}
\toprule
\multirow{2}{*}{\textbf{Method}} & \textbf{Reward Model} & \multicolumn{2}{c}{\textbf{Post-Training}} \\
\cmidrule(lr){2-2} \cmidrule(lr){3-4}
~ & \textbf{ACC} & \textbf{VBench2} & \textbf{MQ}\\
\midrule
BT    & 77.63 & 0.8693
 & 0.1719 \\
\emphcell{BT-WT} & \emphcell{78.27} & \emphcell{0.8999}
 & \emphcell{0.3302} \\
\bottomrule
\end{tabular}
}
\label{tab:bt_vs_btwt}
\end{table}

%% file: table/post_training_performance.tex
\begin{table}[!tb]
    \centering
    \caption{Comparison of reward models on post-training. HunyuanVideo is selected as the video generation backbone, and DanceGRPO algorithm is applied to fine-tune the model. VideoAlign MQ, our reward model score, and VBench2 Human Fidelity are selected as evaluation metrics.
    }
    \adjustbox{width=0.9\linewidth}{
    \begin{tabular}{ll|cccccccc}
         \toprule
         \textbf{Backbone} & \textbf{RM} & \textbf{MQ} & \textbf{{\ours}} & \textbf{VBench2} \\
         \midrule
         HunyuanVideo & - & -0.0980 & 4.5628 & 0.8426 \\
         HunyuanVideo & MQ & 0.1607 & 4.8968 & 0.8695 \\
         \midrule
         \emphcell{HunyuanVideo} & \emphcell{Ours} & \emphcell{0.3302} & \emphcell{5.3554} & \emphcell{0.8999}\\
         \bottomrule
    \end{tabular}
    }
    \label{tab:post_training_performance}
\end{table}

%% file: table/RM-architecture.tex
\begin{table}[!tb]
  \centering
  \caption{Comparison of reward model architecture. $^*$ means score clustering shown in \cref{fig:score_dist_architecture}, where the score distribution degenerates to discrete values.}
  \adjustbox{width=0.9\linewidth}{
  \begin{tabular}{ll|cccccccc}
    \toprule
    \multirow{2}{*}{\textbf{Task}} & \multirow{2}{*}{\textbf{Approach}} & \multicolumn{2}{c}{\textbf{RM ACC}}\\
    \cmidrule(lr){3-4}
    ~ & ~ & \textbf{ID} & \textbf{OOD} \\
    \midrule
    \multirow{4}{*}{Phy \& Deform} & Linear (Ln)~\cite{zhao2025swift} & 74.69 & 78.66 \\
    ~ & `Yes' token logits~\cite{wu2025rewarddance} & 75.43 & 78.46 \\
    ~ & Special token + Ln~\cite{liu2025improving} & 75.91 & 73.61 \\
    ~ & \emphcell{HPQA (Ours)} & \emphcell{78.48} & \emphcell{80.08} \\
    \midrule
    \multirow{4}{*}{TA} & Linear (Ln)~\cite{zhao2025swift} & 72.41$^*$ & 31.92$^*$ \\
    ~ & `Yes' token logits~\cite{wu2025rewarddance} & 71.56$^*$ & 31.37$^*$ \\
    ~ & Special token + Ln~\cite{liu2025improving} & 76.25 & 58.38 \\
    ~ & \emphcell{HPQA (Ours)} & \emphcell{79.02} & \emphcell{60.25} \\
    \bottomrule
  \end{tabular}
  }
  \label{tab:RM-architecture}
\end{table}

%% file: table/RM-BT-BCE.tex
\begin{table}[!tb]
\centering
\caption{
Impact of the BCE Penalty on Reward Model Accuracy and Score Margin. While accuracy is comparable, the BCE penalty significantly reduces the score margin, negatively impacting post-training.
}
\adjustbox{width=0.9\linewidth}{
\begin{tabular}{cc|cc|cc}
\toprule
\multicolumn{2}{c|}{\textbf{Method}} & \multicolumn{2}{c|}{\textbf{Reward Model}} & \multicolumn{2}{c}{\textbf{Post-Training}} \\
\cmidrule(lr){1-2} \cmidrule(lr){3-4} \cmidrule(lr){5-6}
\textbf{BT-WT} & \textbf{BCE} & \textbf{ACC} & \textbf{Margin} & \textbf{VBench2} & \textbf{MQ}\\
\midrule
\checkmark & \checkmark & 78.32 & 2.97 & 0.8826 & 0.1154 \\
\emphcell{\checkmark} & \emphcell{~} & \emphcell{78.48} & \emphcell{3.72} & \emphcell{0.8999} & \emphcell{0.3302} \\
& \checkmark & 75.99 & 1.75 & - & - \\
\bottomrule
\end{tabular}
}
\label{tab:bce_results}
\end{table}

%% file: appendix.tex
\clearpage
\setcounter{page}{1}
\maketitlesupplementary

\section{More Discussions about BT-WT}
\label{sec:whether-pair-win-tie}
While `pass' can form win-ties in single-item annotation, the applicability is dimension-dependent. Binary annotation inherently compresses a continuous quality spectrum into two discrete points.
For some dimensions, `pass' denotes an absolute, discrete state. While for others, `pass' merely signifies surpassing a subjective threshold, masking underlying gradations of quality.
We propose a test to assess a dimension's suitability:
\begin{quote}
Does a shared `pass' label for samples $y_i$ and $y_j$ imply identical underlying quality, or merely a negligible difference?
\end{quote}
\textbf{Yes:} The dimension is suitable for constructing win ties.
\textbf{No} (i.e., $y_i$ may still be superior to $y_j$): The dimension is unsuitable for win ties.

\section{Additonal Experiments}
\subsection{Pairing Strategy}
\label{sec:pairing-strategy}
We ablate the impact of data pairing strategies on reward model performance by comparing two approaches: the in-prompt and cross-prompt pairing strategies. In the in-prompt method, both videos within a preference pair are generated from an identical prompt. Conversely, the cross-prompt strategy permits a pair to be formed from videos generated by different prompts. We construct separate training datasets using each strategy and evaluated the resulting RMs' accuracy and reward margin on fixed evaluation datasets. As shown in \cref{tab:RM-pair_strategy-ACC-ID-OOD} and \cref{tab:RM-pair_strategy-margin-ID-OOD}, the cross-prompt strategy achieves performance comparable to its in-prompt counterpart. Furthermore, a hybrid strategy combining both datasets also yielded similar results. The primary advantage of the cross-prompt approach is its relaxed data requirement: while the in-prompt method necessitates two or more video generations per prompt, the cross-prompt strategy can effectively leverage data from prompts that yielded only a single video. This improves the utilization of available data without compromising reward model performance.

\input{table/RM-pair_strategy-ACC-ID-OOD}
\input{table/RM-pair_strategy-margin-ID-OOD}

\subsection{Model Scaling}
\input{table/RM-model_scale}
As demonstrated in \cref{tab:RM-model_scale}, our analysis of model scaling reveals two key findings. First, performance improvements are substantially more pronounced in the OOD evaluation than in the ID setting. Second, these scaling benefits are not monotonic and are heavily concentrated in the transition from 1B to 8B parameters. This initial scaling step yields significant OOD accuracy gains (up to +5.01) and dramatically improves the OOD reward margin (e.g., from 2.95 to 5.51 for ``Phy \& Deform"), whereas ID accuracy sees only modest increases. In contrast, scaling further from 8B to 14B yields diminishing returns. 

The observed diminishing returns when scaling from 8B to 14B parameters, following significant gains from 1B to 8B, suggest a multi-faceted bottleneck. The initial 1B model is likely under-parameterized, lacking the capacity to capture essential features, which the 8B model successfully acquires. However, the 8B model may already achieve capacity saturation for the given task's intrinsic complexity. This means the additional parameters of the 14B model offer only marginal utility. Furthermore, performance is likely becoming data-limited, where the 14B model requires a larger or more diverse dataset than available to unlock further gains. This is supported by the degradation in the OOD reward margin for both tasks, which indicates the 14B model may be overfitting to the training data, a common challenge when model size outpaces data scale and optimization refinement.

\subsection{BT, BTT and BT-WT}
We evaluate the impact of the BT, Bradley-Terry with Ties~(BTT), and BT-WT loss functions on RM training and subsequent policy optimization. The three losses utilize different data pairing strategies: BT uses only win-lose pairs; BTT incorporates win-lose, win-tie, and lose-tie pairs; and BT-WT employs win-lose and win-tie pairs. The BTT data consists of the BT-WT data plus an additional $150k$ lose-tie pairs.

As discussed in \cref{sec:bt-with-win-tie}, VideoAlign~\cite{liu2025improving} retains all tie samples, using a loss function that models A-wins, B-wins, or ties (which includes both win-tie and lose-tie pairs). We contend this approach is ill-suited for our annotation scenario. Specifically, if two negative samples are labeled independently (e.g., for deformity), we cannot assume their degree of severity is equivalent. Erroneously treating them as a `tie' (a lose-tie pair) would introduce label noise, diminishing the RM's discriminative capacity and ultimately degrading policy performance.

Empirical results in \cref{tab:bt_vs_btt_vs_btwt} support this argument: BTT achieves a lower RM accuracy than BT-WT. Furthermore, the degradation extends to generation quality. As shown in the post-training metrics, the BTT approach results in inferior VBench2 Human Fidelity scores compared to BT-WT. These findings confirm that incorporating lose-tie pairs into the loss function is suboptimal for our data distribution.
\input{table/BT-BTWT-BTT}

\subsection{HPQA Layer Indices}
Empirical results in Table~\ref{tab:hpqa_qwen2} and Table~\ref{tab:hpqa_qwen25_internvl3} demonstrate that the HPQA architecture maintains its efficacy across diverse VLM backbones. Furthermore, our ablation study indicates that aggregating representations from 4 to 5 intermediate layers yields better performance.
\begin{table}[htbp]
  \centering
  \caption{Performance of Qwen2-VL (2B)~\cite{wang2024qwen2} across different layer indices.}
  \label{tab:hpqa_qwen2}
  \renewcommand{\arraystretch}{1.1}
  \adjustbox{max width=\linewidth}{
    \begin{tabular}{ll cccc}
      \toprule
      \multirow{2}{*}{\textbf{Task}} & \multirow{2}{*}{\textbf{Layer Indices}} & \multicolumn{2}{c}{\textbf{ID}} & \multicolumn{2}{c}{\textbf{OOD}} \\
      \cmidrule(lr){3-4} \cmidrule(lr){5-6}
      & & \textbf{ACC} & \textbf{Margin} & \textbf{ACC} & \textbf{Margin} \\
      \midrule
      \multirow{6}{*}{\shortstack[l]{Phy \& \\ Deform}} 
        & `Yes' Token                & 75.40 & 1.71 & 77.05 & 1.69 \\
        & \{14, 28\}                & 75.79 & 1.71 & 77.24 & 1.85 \\
        & \{9, 18, 28\}             & 75.20 & 1.64 & 77.59 & 1.59 \\
        & \{7, 14, 21, 28\}         & \textbf{76.34} & 1.72 & 77.78 & 1.81 \\
        & \{6, 12, 18, 24, 28\}     & 76.29 & 1.77 & \textbf{78.89} & 1.84 \\
        & \{5, 10, 14, 19, 23, 28\} & 74.11 & 1.67 & 75.73 & 1.56 \\
      \midrule
      \multirow{3}{*}{TA} 
        & `Yes' Token                & 77.33 & 3.41 & 55.28 & 0.70 \\
        & \{7, 14, 21, 28\}         & \textbf{78.62} & 3.55 & 56.19 & 0.56 \\
        & \{5, 10, 14, 19, 23, 28\} & 77.53 & 3.17 & \textbf{56.33} & 0.54 \\
      \bottomrule
    \end{tabular}
  }
\end{table}
\begin{table}[htbp]
  \centering
  \caption{Performance of Qwen2.5-VL (3B) and InternVL3 (14B) across different layer indices.}
  \label{tab:hpqa_qwen25_internvl3}
  \renewcommand{\arraystretch}{1.1}
  \adjustbox{max width=\linewidth}{
    \begin{tabular}{ll l cccc}
      \toprule
      \multirow{2}{*}{\textbf{Task}} & \multirow{2}{*}{\textbf{Model}} & \multirow{2}{*}{\textbf{Layer Indices}} & \multicolumn{2}{c}{\textbf{ID}} & \multicolumn{2}{c}{\textbf{OOD}} \\
      \cmidrule(lr){4-5} \cmidrule(lr){6-7}
      & & & \textbf{ACC} & \textbf{Margin} & \textbf{ACC} & \textbf{Margin} \\
      \midrule
      \multirow{7}{*}{\shortstack[l]{Phy \& \\ Deform}} 
        & \multirow{4}{*}{\shortstack[l]{Qwen2.5-VL\\(3B)}} 
          & `Yes' Token                & \textbf{75.45} & 1.61 & 40.76$^*$ & -1.52$^*$ \\
          &                           & \{9, 18, 27, 36\}         & 73.68 & 1.61 & 76.38 & 1.50 \\
          &                           & \{7, 14, 21, 28, 36\}     & 73.51 & 1.47 & \textbf{76.61} & 1.39 \\
          &                           & \{6, 12, 18, 24, 30, 36\} & 72.87 & 1.43 & 74.27 & 1.35 \\
      \cmidrule{2-7}
        & \multirow{3}{*}{\shortstack[l]{InternVL3\\(14B)}} 
          & \{16, 32, 48\}            & 78.10 & 3.56 & 80.99 & 5.39 \\
          &                           & \{12, 24, 36, 48\}        & \textbf{78.50} & 3.52 & 80.31 & 5.17 \\
          &                           & \{9, 18, 27, 36, 48\}     & 78.20 & 3.71 & \textbf{81.16} & 5.21 \\
      \midrule
      \multirow{3}{*}{TA} 
        & \multirow{3}{*}{\shortstack[l]{Qwen2.5-VL\\(3B)}} 
          & `Yes' Token                & 76.19 & 1.89 & 27.73$^*$ & 0.24$^*$ \\
          &                           & \{9, 18, 27, 36\}         & 76.54 & 2.84 & \textbf{59.06} & 0.51 \\
          &                           & \{6, 12, 18, 24, 30, 36\} & \textbf{77.23} & 2.80 & 58.16 & 0.35 \\
      \bottomrule
      \multicolumn{7}{l}{\footnotesize $^*$ scores degenerate to discrete values.}
    \end{tabular}
  }
\end{table}

\subsection{Post-Training}
\subsubsection{Physical Plausibility and Subject Deformity}
We select corresponding VBench2 Human Fidelity, including human anatomy, human clothes and human identity, as the evaluation dimensions. As detailed in ~\cref{tab:post_train_vbench2_detail}, our RM outperforms both the baseline and the VideoAlign-MQ guided training.
Both BT-WT and HPQA generate positive returns, with the former yielding even higher returns.
\input{table/post_training_performance_detail}

\subsubsection{Semantic Alignment}
\label{sec:post_training_TA}
\cref{tab:vbench1_semantic_abbrev} summarizes the performance comparison of the HunyuanVideo backbone after post-training with different reward models, evaluated on the VBench benchmark for semantic alignment. Both post-training methods, using the VideoAlign TA reward model and our {\ours}, improve upon the baseline HunyuanVideo model's semantic score of 0.7334. Our proposed reward model achieves the highest overall performance, with a semantic score of \textbf{0.7544}, surpassing the TA model's score of 0.7421. 
\input{table/post_train_TA}

\section{Data Annotation}
\label{appendix:sec_annotation}
\subsection{Single-Item Binary Annotation Design}
To ensure a standardized and rigorous evaluation process, we formulate detailed annotation guidelines for our human annotators. These guidelines, summarized in \cref{tab:evaluation_dimensions_styled}, provide a structured framework for assessing the quality of generated videos. The evaluation is organized into three primary dimensions. The first dimension, \textbf{Subject Deformity}, focuses on the structural plausibility and temporal stability of the subjects within the video. The second dimension, \textbf{Physical Plausibility}, evaluates the adherence of the video's dynamics to real-world physical laws, including motion, gravity, and object interactions. The final dimension, \textbf{Semantic Alignment}, measures the relevance of the generated video content to the input text prompt, encompassing core semantics, detailed descriptions, and stylistic specifications. 
\input{table/annotation_rule}
To simplify the annotation task and ensure consistency, annotators are instructed to perform a binary assessment for each dimension. They are required solely to judge whether a video passes or fails the specific criteria defined for that dimension.

\subsection{Data Distribution}
\begin{figure}[!tb]
    \centering
    \includegraphics[width=0.8\linewidth]{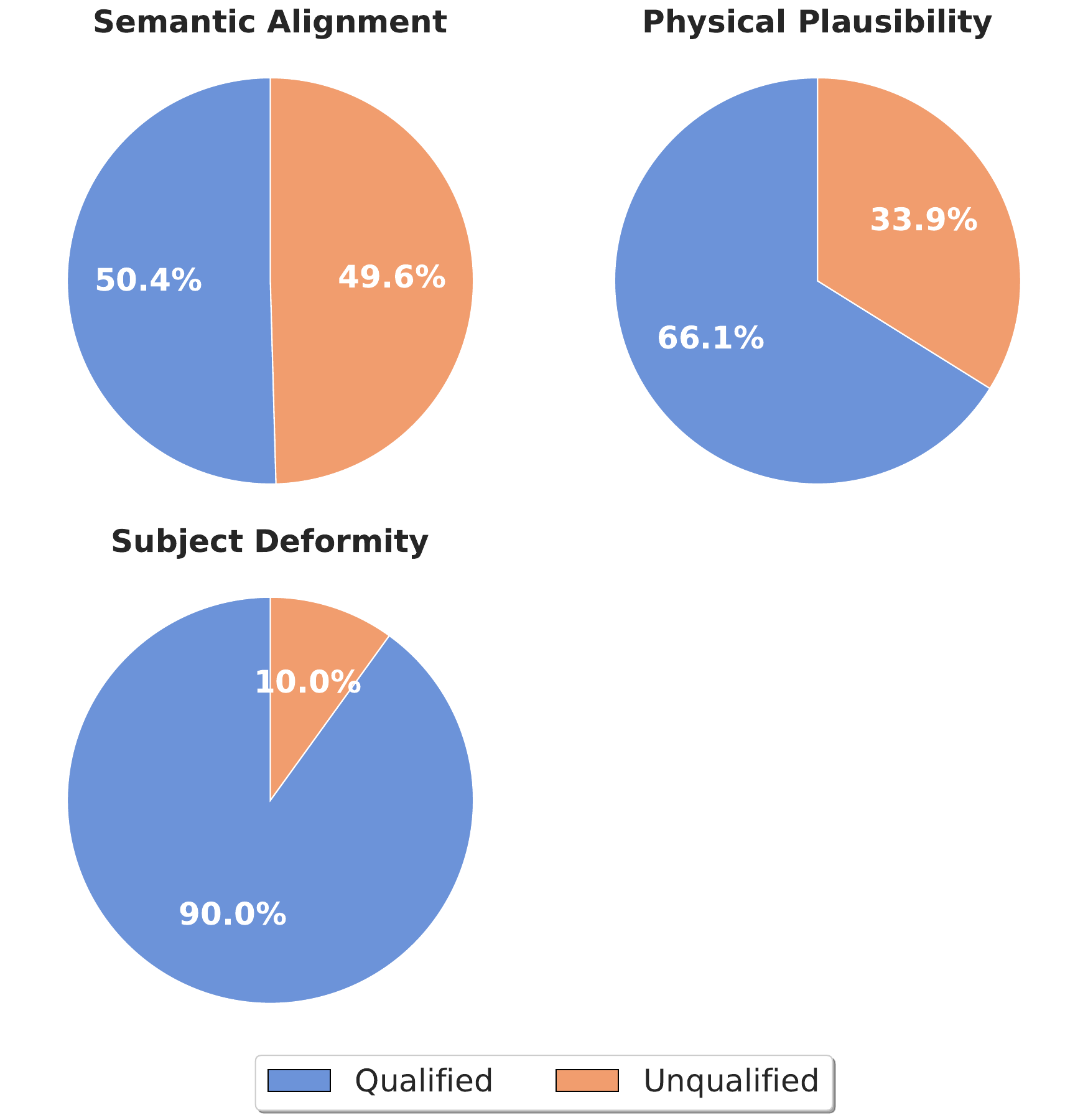}
    \caption{Quality distribution of our in-house dataset.}
    \label{fig:data-distribution}
\end{figure}

We collected annotations for $250\mathrm{k}$ in-house videos and a $50\mathrm{k}$-sample out-of-distribution (OOD) test set generated by other SOTA models~(including Wan2.1~\cite{wan2025wan}, Wan2.2~\cite{wan2025wan}, Veo 3~\cite{google2025veo} and Seedance 1.0~\cite{gao2025seedance}). These samples originate from $20\mathrm{k}$ unique prompts. To ensure diversity, we balanced these prompts across multiple attributes including subject, motion, style, and camera control.
In \cref{fig:data-distribution}, we demonstrate the quality distribution of our in-house dataset.
To construct the BT-WT dataset, we randomly selected 350$k$ win-lose and 150$k$ win-tie pairs based on the cross-prompt pairing strategy.

\section{Implementation Details}
\subsection{Reward Model Training}
We trained the reward model on a 4-node CentOS cluster, each node featuring 8 NVIDIA H20 GPUs with 96GB CUDA memory and an AMD EPYC 9K84 96-Core Processor. All other hyperparameters are detailed in \cref{tab:imple_details-RM}.

\subsection{Post-Training}
We conducted post-training experiments on a 5-node CentOS cluster, each node featuring 8 NVIDIA H800 GPUs (80GB CUDA memory) and an Intel Xeon Platinum 8476C Processor. Key hyperparameters are detailed in \cref{tab:imple_details-post_train}.

\section{More Visualization Results}
\begin{figure*}[!htbp]
    \centering
    \includegraphics[width=1.0\linewidth]{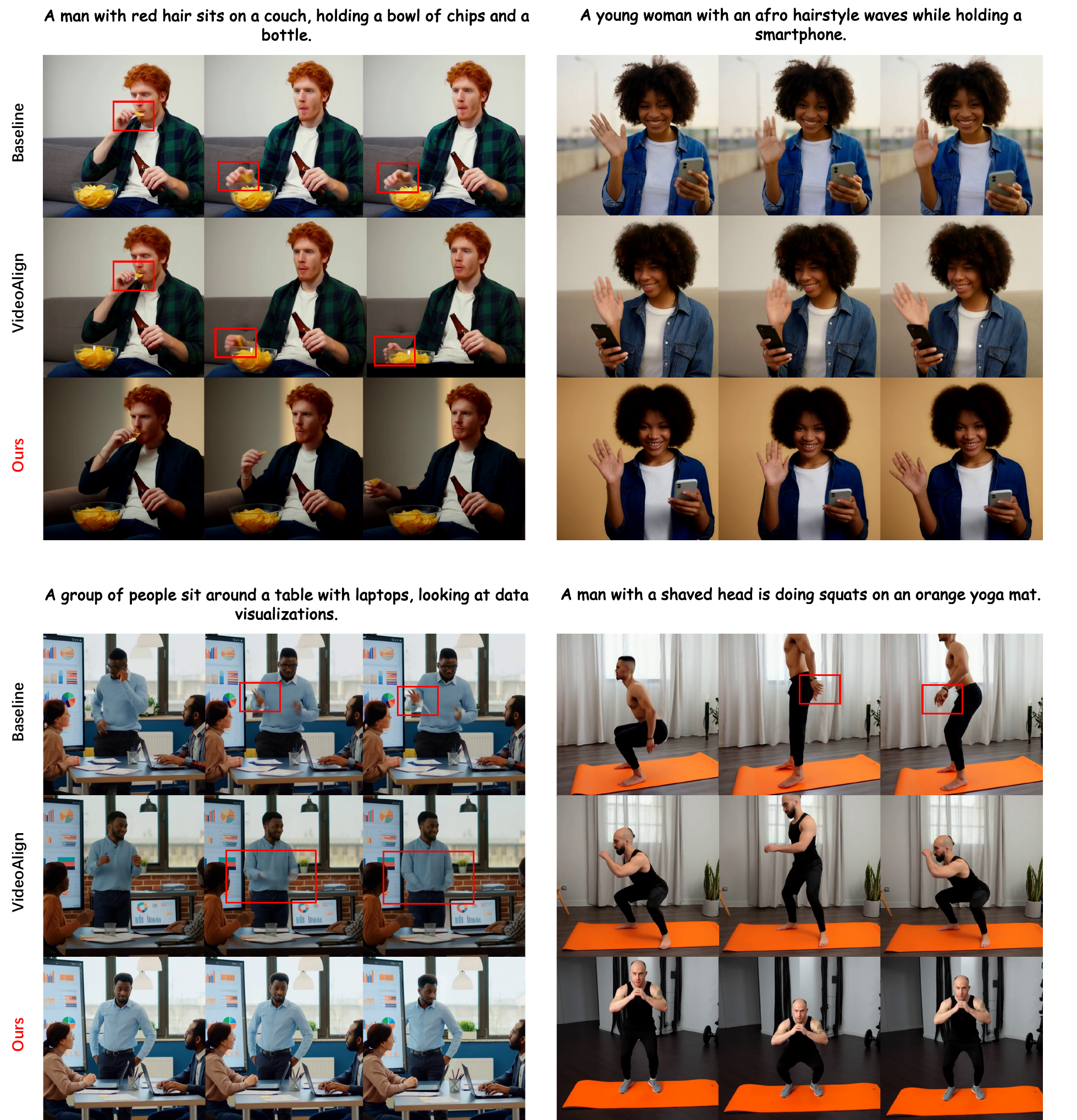}
    \caption{More visualization results guided by VideoAlign and \ours.}
    \vspace{-12pt}
    \label{fig:phy_deformity_case}
\end{figure*}

\begin{figure*}[!htbp]
    \centering
    \includegraphics[width=1.0\linewidth]{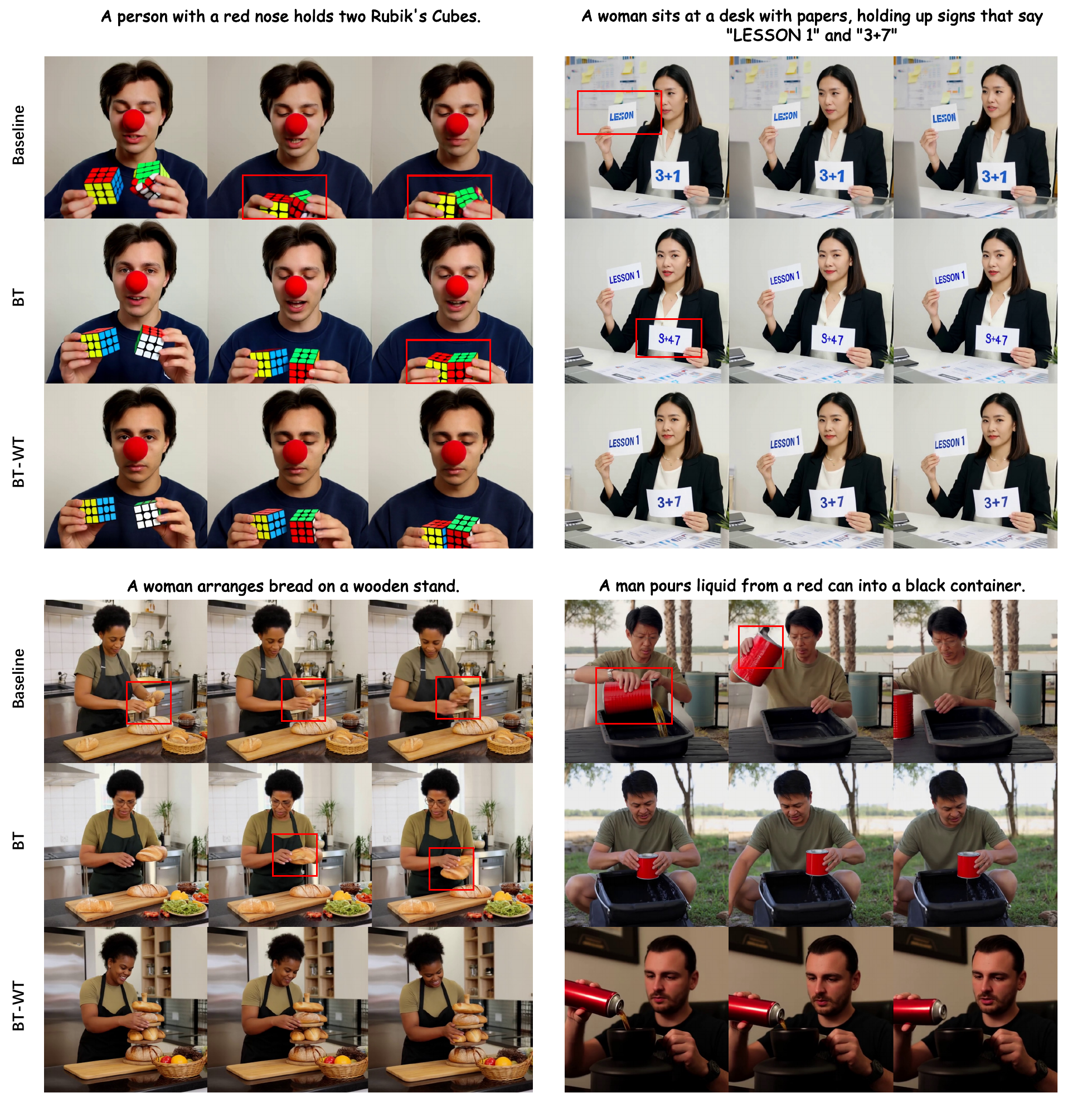}
    \caption{Visualization results guided by BT and BT-WT.}
    \vspace{-12pt}
    \label{fig:phy_deformity_case_bt_btwt_case}

\end{figure*}

Further visualization results are presented in \cref{fig:phy_deformity_case}.
We also provide visual comparisons between BT and BT-WT in \cref{fig:phy_deformity_case_bt_btwt_case}.
For extensive qualitative results, please refer to the supplementary material, which includes HTML pages for convenient viewing.

\begin{table*}[!htbp] 
  \centering
  \begin{minipage}[t]{0.48\linewidth}
    \centering
    \caption{Implementation details for reward model training.}
    \label{tab:imple_details-RM}
    \adjustbox{max width=\linewidth}{
      \begin{tabular}{ll}
        \toprule
        \textbf{Hyperparameter} & \textbf{Value} \\
        \midrule
        Model & InternVL3-1B/8B/14B \\
        Distributed Training & DeepSpeed Stage 0/3/3 \\
        Trainable Parameters & Full \\
        Learning Rate & 1e-6 \\
        Num. Train Epochs & 3.0 \\
        Per Device Train Batch Size & 1 \\
        Gradient Accumulation Steps & 10 \\
        Optimizer & AdamW \\
        Adam $\beta_1$ & 0.9 \\
        Adam $\beta_2$ & 0.999 \\
        Weight Decay & 0.01 \\
        LR Scheduler & linear \\
        Warmup Ratio & 0.05 \\
        Reward Margin & 3.0 \\
        Precision & BF16 \\
        Gradient Checkpointing & True \\
        \bottomrule
      \end{tabular}
    }
  \end{minipage}%
  \hfill 
  \begin{minipage}[t]{0.48\linewidth}
    \centering
    \caption{Implementation details for post-training.}
    \label{tab:imple_details-post_train}
    \adjustbox{max width=\linewidth}{
      \begin{tabular}{ll}
        \toprule
        \textbf{Hyperparameter} & \textbf{Value} \\
        \midrule
        Base Model & HunyuanVideo 14B \\
        Trainable Parameters & Full \\
        Learning Rate & 1e-6 \\
        Per Device Train Batch Size & 1 \\
        Gradient Accumulation Steps & 4 \\
        Max Train Steps & 500 \\
        LR Scheduler & constant with warmup \\
        LR Warmup Steps & 0 \\
        Mixed Precision & bf16 \\
        Gradient Checkpointing & True \\
        SDE $\eta$ & 0.25 \\
        Video Resolution (train) & $480 \times 480$ \\
        Video Frames (train) & 32 \\
        Video FPS (train) & 8 \\
        Denoising Steps (train) & 16 \\
        Time Shift (train) & 5.0 \\
        Video Resolution (test) & $640 \times 640$ \\
        Video Frames (test) & 91 \\
        Video FPS (test) & 18 \\
        Denoising Steps (test) & 30 \\
        Time Shift (test) & 7.0 \\
        Group Size & 8 \\
        Distributed Training & FSDP FULL SHARD \\
        \bottomrule
      \end{tabular}
    }
  \end{minipage}
\end{table*}

%% file: table/RM-pair_strategy-ACC-ID-OOD.tex
\begin{table}[!tb]
 \centering
 \caption{Impact of pairing-strategy for reward model training. Accuracy is evaluated on both ID and OOD datasets. The cross-prompt strategy is comparable with in-prompt strategy.
 }
 \adjustbox{width=0.9\linewidth}{
 \begin{tabular}{ll|cc}
  \toprule
  \multirow{2}{*}{\textbf{Task}} & \multirow{2}{*}{\textbf{Approach}} & \multicolumn{2}{c}{\textbf{Reward ACC}} \\
  \cmidrule(lr){3-4} 
  & & \textbf{ID} & \textbf{OOD} \\
  \midrule
  \multirow{3}{*}{Phy \& Deform} & Cross-Prompt & 76.74 & 79.54 \\
  ~ & In-Prompt & 76.77 & 79.22 \\
  ~ & Hybrid & 76.09 & 79.16 \\
  \midrule
  \multirow{3}{*}{TA} & Cross-Prompt & 76.39 & 60.25 \\
  ~ & In-Prompt & 76.67 & 59.26 \\
  ~ & Hybrid & 75.64 & 59.41 \\
  \bottomrule
 \end{tabular}
 }
 \label{tab:RM-pair_strategy-ACC-ID-OOD}
\end{table}

%% file: table/RM-pair_strategy-margin-ID-OOD.tex
\begin{table}[!tb]
 \centering
 \caption{Impact of pairing-strategy for reward model training. Pair score margin is evaluated on both ID and OOD datasets. The cross-prompt strategy is comparable with in-prompt strategy.
 }
 \adjustbox{width=0.9\linewidth}{
 \begin{tabular}{ll|cc}
  \toprule
  \multirow{2}{*}{\textbf{Task}} & \multirow{2}{*}{\textbf{Approach}} & \multicolumn{2}{c}{\textbf{Reward Margin}} \\
  \cmidrule(lr){3-4}
  & & \textbf{ID} & \textbf{OOD} \\
  \midrule
  \multirow{3}{*}{Phy \& Deform} & Cross-Prompt & 3.83 & 4.17 \\
  ~ & In-Prompt & 3.98 & 3.77 \\
  ~ & Hybrid & 3.98 & 4.13 \\
  \midrule
  \multirow{3}{*}{TA} & Cross-Prompt & 2.93 & 1.47 \\
  ~ & In-Prompt & 2.73 & 1.28 \\
  ~ & Hybrid & 2.87 & 1.37 \\
  \bottomrule
 \end{tabular}
 }
 \label{tab:RM-pair_strategy-margin-ID-OOD}
\end{table}

%% file: table/RM-model_scale.tex
\begin{table}[!tb]
  \centering
  \caption{Influence of model size on reward model accuracy and reward margin between positive and negative samples.}
  \adjustbox{width=0.9\linewidth}{
  \begin{tabular}{ll|cccccccc}
    \toprule
    \multirow{2}{*}{\textbf{Task}} & \multirow{2}{*}{\textbf{Approach}} & \multicolumn{2}{c}{\textbf{ID}} & \multicolumn{2}{c}{\textbf{OOD}} \\
    \cmidrule(lr){3-4} \cmidrule(lr){5-6}
    ~ & ~ & \textbf{ACC} & \textbf{Margin} & \textbf{ACC} & \textbf{Margin} \\
    \midrule
    \multirow{3}{*}{Phy \& Deform} & InternVL3-1B & 76.25 & 3.10 & 77.65 & 2.95 \\
    ~ & InternVL3-8B & 78.48 & 3.60 & 81.43 & 5.51 \\
    ~ & InternVL3-14B & 78.57 & 3.69 & 81.71 & 5.19 \\
    \midrule
    \multirow{3}{*}{TA} & InternVL3-1B & 77.48 & 2.08 & 58.45 & 0.33 \\
    ~ & InternVL3-8B & 79.02 & 2.11 & 63.46 & 0.92 \\
    ~ & InternVL3-14B & 78.83 & 2.15 & 65.29 & 0.71 \\
    \bottomrule
  \end{tabular}
  }
  \label{tab:RM-model_scale}
\end{table}

%% file: table/BT-BTWT-BTT.tex
\begin{table}[!tb]
\centering
\caption{Comparison of reward model performance trained via BT, BTT and BT-WT. Reward model accuracy and VBench2 Human Fidelity scores are reported.
}
\adjustbox{width=0.8\linewidth}{
\begin{tabular}{l|c|cccc}
\toprule
\multirow{2}{*}{\textbf{Method}} & \textbf{Reward Model} & \multicolumn{2}{c}{\textbf{Post-Training}} \\
\cmidrule(lr){2-2} \cmidrule(lr){3-4}
~ & \textbf{ACC} & \textbf{VBench2} & \textbf{MQ}\\
\midrule
BT    & 77.63 & 0.8693 & 0.1719 \\
BTT    & 77.78 & 0.8700 & 0.0690 \\
\emphcell{BT-WT} & \emphcell{78.27} & \emphcell{0.8999} & \emphcell{0.3302} \\
\bottomrule
\end{tabular}
}
\label{tab:bt_vs_btt_vs_btwt}
\end{table}

%% file: table/post_training_performance_detail.tex
\begin{table}[!tb]
  \centering
  \caption{Comparison of different approaches on human preference benchmarks from VBench2~\cite{zheng2025vbench}. The evaluation metrics include Anatomy (Human Anatomy), Clothes (Human Clothes), Identity (Human Identity), and Fidelity (Human Fidelity).}
  \label{tab:post_train_vbench2_detail}
  \renewcommand{\arraystretch}{1.1} 
  \adjustbox{max width=\linewidth}{ 
    \begin{tabular}{l cccc}
      \toprule
      \textbf{Method} & \textbf{Anatomy} & \textbf{Clothes} & \textbf{Identity} & \textbf{Fidelity (Overall)} \\
      \midrule
      Baseline & 0.8915 & 0.8905 & 0.7457 & 0.8426 \\
      VideoAlign-MQ & 0.9009 & 0.8714 & 0.8361 & 0.8695 \\
      \midrule
      `Yes' Token w/ BT & 0.9154 & 0.9067 & 0.7778 & 0.8666 \\
      `Yes' Token w/ BT-WT & \textbf{0.9312} & \textbf{0.9259} & 0.8013 & 0.8861 \\
      HPQA w/ BT & 0.9153 & 0.9064 & 0.7861 & 0.8693 \\
      HPQA w/ BTT & 0.9254 & 0.8922 & 0.7923 & 0.8700 \\
      HPQA w/ BT-WT & 0.9164 & 0.9231 & \textbf{0.8601} & \textbf{0.8999} \\
      \bottomrule
    \end{tabular}
  }
\end{table}


%% file: table/post_train_TA.tex
\begin{table*}[!tb]
\vspace{-5\baselineskip}
\centering
\caption{Comparison of post-training with different reward models on semantic alignment. Evaluations are conducted on VBench~\cite{huang2024vbench}.}

\adjustbox{width=\textwidth}{
\begin{tabular}{ll|ccccccccc|c}
\toprule
\textbf{Backbone} & \textbf{RM} & 
\textbf{Scene} & 
\textbf{Consistency} & 
\textbf{Appearance} & 
\textbf{Object} & 
\textbf{Spatial} & 
\textbf{Action} & 
\textbf{Temporal} & 
\textbf{Color} & 
\textbf{Multiple} & 
\textbf{Semantic (Avg.)} \\
\midrule
HunyuanVideo & - & 0.3496 & 0.2700 & 0.2021 & 0.8006 & 0.6993 & 0.9600 & \emphcell{0.2539} & 0.8730 & 0.6966 & 0.7334 \\

HunyuanVideo & TA & \emphcell{0.3895} & \emphcell{0.2702} & \emphcell{0.2039} & 0.7927 & 0.7150 & \emphcell{0.9700} & 0.2539 & 0.8269 & 0.7477 & 0.7421 \\

HunyuanVideo & Ours & 0.3692 & 0.2678 & 0.1981 & \emphcell{0.8275} & \emphcell{0.7517} & \emphcell{0.9700} & 0.2518 & \emphcell{0.9080} & \emphcell{0.7630} & \emphcell{0.7544} \\
\bottomrule
\end{tabular}
}
\label{tab:vbench1_semantic_abbrev}
\end{table*}

%% file: table/annotation_rule.tex
\begin{table*}[!htbp]
    \centering
    \caption{Definitions of the evaluation dimensions and their key assessment criteria used in our human evaluation framework.}
    \label{tab:evaluation_dimensions_styled}
    \begin{tabular}{c|c}
        \toprule
        \textbf{Evaluation Dimension} & \textbf{Definition and Key Criteria}  \\
        \midrule
        \textbf{Subject Deformity} & \begin{tabular}{p{0.7\textwidth}}
        Assesses the presence and severity of \textbf{structural artifacts and temporal instability} impacting subjects (e.g., humans, animals, objects).
        
        \begin{itemize}[label={-}, leftmargin=15pt, topsep=3pt, partopsep=0pt, itemsep=2pt]
            \item \textbf{Structural Artifacts:} Evaluates anatomical incorrectness, penalizing severe distortions, unnatural forms, or implausible subject parts (e.g., faces, limbs).
            \item \textbf{Temporal Instability:} Measures inconsistencies in a subject's identity or form, penalizing artifacts like ``melting", ``flickering", or unnatural morphing across frames.
        \end{itemize}
        \end{tabular} \\
        \midrule
        \textbf{Physical Plausibility} & \begin{tabular}{p{0.7\textwidth}}
        Assesses the adherence of video dynamics to real-world \textbf{physical principles}.
        
        \begin{itemize}[label={-}, leftmargin=15pt, topsep=3pt, partopsep=0pt, itemsep=2pt]
            \item \textbf{Motion Dynamics:} Checks if object motion, acceleration, and inertia appear physically coherent and natural.
            \item \textbf{Object Interactions:} Evaluates the realism of interactions with forces like gravity (e.g., falling) and between entities (e.g., collisions, splashes).
            \item \textbf{Material Dynamics:} Assesses the realistic behavior and deformation of complex materials such as fluids (water, smoke) or soft bodies (cloth).
        \end{itemize}
        \end{tabular} \\
        \midrule
        \textbf{Semantic Alignment} & \begin{tabular}{p{0.7\textwidth}}
        Assesses the \textbf{fidelity} of the generated video content with respect to the input text prompt.
        
        \begin{itemize}[label={-}, leftmargin=15pt, topsep=3pt, partopsep=0pt, itemsep=2pt]
            \item \textbf{Core Semantics:} Fidelity to the primary semantic components of the prompt, including the main subject, key action, and overall scene.
            \item \textbf{Detailed Attributes:} Fidelity to specific descriptive details, such as subject attributes (e.g., color, appearance), action modifiers, and background elements.
            \item \textbf{Stylistic \& Cinematic Fidelity:} Alignment with specified artistic styles (e.g., 3D render) and cinematic instructions (e.g., camera motion, ``close-up").
        \end{itemize}
        \end{tabular} \\
        \bottomrule
    \end{tabular}
\end{table*}